\pdfoutput=1 

\documentclass[11pt]{article}
\usepackage[]{emnlp2021}
\usepackage{nert}

\usepackage{times}
\usepackage{latexsym}
\usepackage{linguex}
\usepackage{caption}
\usepackage{subcaption}
\usepackage{changepage}
\usepackage{microtype}

\title{BERT Has Uncommon Sense:\\ Similarity Ranking for Word Sense BERTology}

\author{Luke Gessler \quad Nathan Schneider \\ 
  Georgetown University \\
  {\{\emldisplay{lg876@georgetown.edu}{lg876},
  \emldisplay{nathan.schneider@georgetown.edu}{nathan.schneider}\}\texttt{@georgetown.edu} }}

\date{}

\begin{document}
\maketitle
\begin{abstract}
An important question concerning contextualized word embedding (CWE) models like BERT is how well they can represent different word senses, especially those in the long tail of uncommon senses. 
Rather than build a WSD system as in previous work, we investigate contextualized embedding neighborhoods directly, formulating a query-by-example nearest neighbor retrieval task and examining ranking performance for words and senses in different frequency bands. 
In an evaluation on two English sense-annotated corpora, we find that several popular CWE models all outperform a random baseline even for proportionally rare senses, without explicit sense supervision. 
However, performance varies considerably even among models with similar architectures and pretraining regimes, with especially large differences for rare word senses, revealing that CWE models are not all created equal when it comes to approximating word senses in their native representations. 
\end{abstract}






\section{Introduction}

Contextualized word embedding (CWE) models such as BERT \citep{devlin_bert_2019}, which were enabled by Transformers \citep{vaswani_attention_2017}, have yielded great improvements in a variety of NLP tasks.
However, because BERT and other CWE models are deep neural networks with complicated architectures and very high parameter counts, it is not easy to understand exactly which aspects of linguistic form and meaning contextualized word embeddings are able to capture.

In response to these difficulties, much work has been done attempting to interpret CWE models, most notably in the field of BERTology. 
Over 100 BERTological studies have been published since BERT was introduced in 2018, covering a diverse range of linguistic phenomena \citep{rogers_primer_2020} such as POS tags, constituency, and event factuality \citep{liu_linguistic_2019}.
The methods developed in this area are usually applicable to other CWE models.

An important question for CWE models is how well they can represent rare word senses.
Word sense disambiguation systems have been observed to be most lacking in the long tail of rare word senses \citep{blevins_moving_2020, blevins_fews_2021}, and sense-awareness is of fundamental importance for many NLP applications.
Thus a better understanding of how well CWE models understand senses, and especially rare senses, would aid the interpretation of NLP systems.

To address this question, we perform an evaluation we call CWE similarity ranking on two sense-annotated English corpora using several popular CWE models.
We find that while all models outperform a random baseline on the evaluation, models differ substantially in their performance on rare word senses, with significant differentiation even between models which are closely related.%
\footnote{All code for this work is available at \url{https://github.com/lgessler/bert-has-uncommon-sense}.}

\section{Previous Work}
\subsection{Word Senses}
The task of word sense disambiguation (WSD) introduced the notion of a \textit{word sense} into NLP \citep{navigli_word_2009,ijcai2021-593}. 
The WSD task is typically formulated as labeling words in context with their senses as defined by a dictionary or other lexical resource.

Many resources exist to support work on word senses.
WordNet \citep{miller_wordnet_1995} has been a crucial resource for work on word senses, providing a fine-grained and comprehensive inventory of words and their senses for English.
Several large annotated corpora have been constructed using WordNet senses, including SemCor \citep{semcor,landes_building_1998} and OntoNotes \citep{hovy_ontonotes_2006}.
More recently, WSD systems have been able to achieve human-like performance on WSD tasks, but performance on rare senses remains comparatively poor \citep{blevins_moving_2020}, leading to the construction of corpora tailored for assessing systems on rare senses \citep{blevins_fews_2021}.

\subsection{Word Sense BERTology} 
\label{sec:wsb}
Few studies have focused narrowly on the question of how well BERT and other CWE models capture word senses. 
In one such study, \citet{wiedemann_does_2019} evaluate CWE models by using the CWE models' embeddings as representations for a kNN classifier: the predicted sense of a word is the one most represented among by the word's $k$ nearest neighbors (via cosine distance).
Despite the simplicity of this WSD system, it is able to achieve results that are competitive with state-of-the-art systems, even achieving new SOTA scores on the SensEval-2 and SensEval-3 datasets.
\citet{reif-2019-geometry} construct a similar WSD system, relying not on kNN but closest sense-centroids (centroids are constructed using a labeled training set) to decide on a label.

Other studies have approached the question by modifying BERT's training scheme.
\citet{tayyar_madabushi_cxgbert_2020} train a BERT variant where the next sentence prediction task has been replaced with a same construction prediction task and find mixed results on downstream tasks. 
\citet{levine_sensebert_2020} train a BERT variant by adding a new supersense prediction task, wherein the masked token's WordNet supersense is to be predicted, and find performance gains on a variety of meaning-related tasks, which shows that BERT's representations do not perfectly capture word senses.

Some recent approaches have developed specialized methods for exploring CWE models' embedding spaces. 
\citet{karidi-bert-2021} present an iterative method for surveying the ``topography'' of BERT's embedding space, finding that word senses often form cohesive regions.

\section{CWE Similarity Ranking}
\label{sec:cwesr}

\begin{figure*}
    \centering
    
    \begin{subfigure}[h]{0.49\textwidth}
        \small
        \begin{adjustwidth}{10pt}{10pt}
        {\bf Query}: But with all the money and glamour of high finance come the relentless pressures to do well; pressure to {\bf pull}$_4$ off another million before lunch [...]
        
        {\bf 1}: ``Sometimes,'' he says, ``we'll {\bf pull}$_{\color{red} 3}$ someone off phones for more training.''
        
        {\bf 2}: Hence, they have never lacked their own stately or amusing charms to {\bf pull}$_{\color{red} 2}$ in wealth and keep it within a household.
        
        {\bf 3}: I can't {\bf pull}$_4$ it off.
        
        {\bf 4}: Bulatovic says Kostunica was able to {\bf pull}$_4$ off the balancing act because he is not really anti-American.
        \end{adjustwidth}
        
        \caption{A sample of a single query on OntoNotes for the verb {\it pull}. Sense 4, glossed as `commit, do', is rare, comprising just 5 of the 78 occurrences of {\it pull} in the OntoNotes training set. Here, the first two results are incorrect: sense 3 means `eliminate from a situation', and sense 2 means `steer something in a certain direction'. All 5 correct instances are retrieved in the top 50 results, at ranks 3, 4, 5, 6, and 41.}
        \label{fig:query_sample_v}
    \end{subfigure}
    \hfill
    \begin{subfigure}[h]{0.49\textwidth}
        \small\centering
        
        %
        %
        %
        %
        %
        \begin{adjustwidth}{10pt}{10pt}
        {\bf Query}: These days I take five pills a day, but at one point I was {\bf on}$_{20}$ about 20.
        
        {\bf 1}: I'm happier than I was three years ago, when I was drinking and I was {\bf on}$_{20}$ cocaine.
        
        {\bf 2}: I was {\bf on}$_{20}$ about sixty cigarettes a day.
        
        {\bf 3}: I was {\bf on}$_{20}$ heavy duty painkillers for 48 hours.
        
        {\bf 4}: He had been put {\bf on}$_{20}$ prescription drugs to help him cope with coming off crack.
        
        {\bf 5}: Recorded in 21 days in a Mitcham garage, the fact that it is {\bf on}$_{\color{red} 2}$ Chrysalis is a mere coincidence.
        
        {\bf 6}: You can phone now {\bf on}$_{\color{red} 12}$ o-five-hundred , four-o-four , treble zero to put your views to Tory MP Phil Gallie .
        \end{adjustwidth}
        
        \caption{A sample of a single query on PDEP for the preposition {\it on}. Sense 20, glossed as `regularly taking (a drug or medicine)', is very rare, comprising 14 of the 1728 occurrences of {\it on} in the PDEP training set. The 5th and 6th results are incorrect: sense 2 has to do with location, and sense 12 a medium of communication. Only 6 of the 14 total occurrences in $\mathcal{D}$ make it into the top 50 for this query.}
        \label{fig:query_sample_p}
    \end{subfigure}
    
    \caption{Query samples using {\tt\small bert-base-cased}.}
    \label{fig:query_sample}
\end{figure*}

We formulate a query-by-example task in which a word in sentence context is used to query for similar usages of the same word in other sentences.
In a process we call \textbf{contextualized word embedding (CWE) similarity ranking}, we will assume some embedding function $f$ and two corpora of sense-labeled text segmented into sentences: a larger ``database'' corpus $\mathcal{D}$, and a smaller ``query'' corpus $\mathcal{Q}$.
We will use sentences from $\mathcal{Q}$ to rank sentences in $\mathcal{D}$, as detailed below.

\begin{enumerate}
    \item Select a query sentence from $\mathcal{Q}$ consisting of tokens $w^{(q)}_{1}, \ldots, w^{(q)}_{n}$, where a token $w^{(q)}_i$ has been designated as the {\it target token} and has sense $s_i$.
    
    \item Find embeddings for the query sentence, $\mathbf{h}^{(q)}_1, \ldots, \mathbf{h}^{(q)}_n = f\left(w^{(q)}_1, \ldots, w^{(q)}_n\right)$.
    
    
    \item For every instance $w_j^{(d)}$  
    with its context $w^{(d)}_1, \ldots, w^{(d)}_m$ in $\mathcal{D}$, and sense $s_j$, find embeddings $\mathbf{h}^{(d)}_1, \ldots, \mathbf{h}^{(d)}_m = f\left(w^{(d)}_1, \ldots, w^{(d)}_m\right)$.
    
    
    \item Rank instances in $\mathcal{D}$ in descending order by cosine similarity between embeddings of the two tokens, $w_i^{(q)}$ and $w_j^{(d)}$: 
    $\textsc{cos-sim}\left(\mathbf{h}^{(q)}_{i}, \mathbf{h}^{(d)}_j\right)$.
    
    \item Evaluate the ranking, e.g.~with precision at $k$.
\end{enumerate}

CWE similarity ranking consists of much of the same work that a CWE-based kNN system would do,%
\footnote{Indeed, the basic method of finding embeddings that are nearest to a target embedding has been widely used both before and after the rise of CWEs (cf. \citealt{wiedemann_does_2019}, which we describe in \Cref{sec:wsb}, and \citealt{schnabel_evaluation_2015}).}
with the difference that the kNN WSD evaluation will only reward a model if the gold sense is held by a plurality of the neighbors, whereas CWE similarity ranking is less stringent and will award ``partial credit'' for retrieving any gold instances, even if they do not form a majority.
This provides a clearer view of how coherent rare word senses are in CWEs' representations.
See \Cref{fig:query_sample} for query examples.

%
%

\section{Experimental Setup}

We use the CWE similarity ranking method to evaluate several popular pretrained CWE models retrieved from \url{huggingface.co} \citep{wolf_huggingfaces_2020}.

\paragraph{Corpora}
Our two corpora are OntoNotes 5.0 \citep{hovy_ontonotes_2006}, which has sense annotations for nouns and verbs,\footnote{Specifically, we use the OntoNotes English noun and verb \emph{sense groupings}, whose sense inventory was formulated by merging WordNet senses for each lemma until an acceptable level of interannotator agreement was reached. Our preliminary experiments with WordNet senses in another corpus, SemCor \citep{langone_annotating_2004}, were difficult to interpret because annotations often seemed to be inconsistent.
} 
and the Pattern Dictionary of English Prepositions (PDEP) corpus \citep{litkowski-2014-pattern}, which has sense annotations for prepositions.\footnote{We use a copy of PDEP obtained from a SQL dump dated 2019-04-19. Original data is available with our code.}
For OntoNotes, we discard instances labeled with ``none-of-the-above'' senses, as we expect them to be heterogeneous. 
For PDEP, we use only instances of 48 common English prepositions. 
For both corpora, we use only single-word targets, and use pre-existing train--validation--test splits.
We treat the training split of each as our $\mathcal{D}$, and the combined validation and test split of each as our $\mathcal{Q}$, and discard any senses in $\mathcal{Q}$ that did not occur at least 5 times in $\mathcal{D}$, as these senses are so rare in $\mathcal{D}$ that performance on them is liable to be overtaken by noise.
For OntoNotes, $|\mathcal{D}|=\text{229,989}$, $|\mathcal{Q}|=\text{50,395}$. For PDEP, $|\mathcal{D}|=\text{33,090}$, $|\mathcal{Q}|=\text{8,020}$.

\paragraph{Inoculation by Fine-Tuning}
In addition to the base models, we also evaluate versions of the models that have been fine-tuned with a small number of instances from another dataset, a method called {\it inoculation by fine-tuning} \citep{richardson_probing_2019,liu_inoculation_2019}.
Inoculation by fine-tuning allows model to surface more domain-relevant information in its output embeddings, while using only a small amount of data so as to avoid teaching the model anything entirely new. 
We use STREUSLE 4.4 \citep{schneider_corpus_2015,schneider_comprehensive_2018} to fine-tune, sampling supersense annotations of single-word nouns, verbs and prepositions in equal numbers for total counts of 100, 250, 500, 1000, and 2500.
See \Cref{sec:ft} for full details.

\paragraph{Layer Choice} 
In this work, we use embeddings from the last layer of every model we assess. 
Preliminary experiments showed that many models show no improvement past the middle layers, and using the last layer is also of interest because many systems freeze their CWE models' weights and use embeddings from their final layers.

\paragraph{Lemma Restriction}
In all cases, neighbors are restricted to instances that have the same lemma.

\paragraph{Evaluation}
To assess the quality of a ranking, we use mean average precision\footnote{Recall is omitted here---see \Cref{sec:results_apdx}.} for the top 50 ranked instances.%
\footnote{Average precision is defined for a single query as the average of precision calculated for every result from 1 to some $k$, where $k$ ranges from 1 to a maximum of 50 for the present study. Mean average precision in turn is the average of this quantity across all instances in $\mathcal{Q}$.}
Since there can be very few gold-labeled instances, it may be impossible for a model to achieve a perfect score.
To make these metrics more interpretable for a given dataset, we also include a baseline score obtained by randomly ranking results, and an oracle upper bound, i.e.~the score that would be obtained by a perfect model.\footnote{Note that this is not always 100\%: for queries with fewer than 50 gold instances that can be retrieved in all of $\mathcal{D}$, the oracle's performance will be under 100\%.}

\section{Results}

\begin{table*}[h!]
    \setlength{\tabcolsep}{2.5pt}
    \renewcommand{\arraystretch}{0.8}
    \begin{subtable}[h]{0.49\textwidth}
        \centering
        \begin{tabular}{c|rrrr}
        {\small Model} & \begin{tabular}{@{}c@{}}{\small $\ell<500$} \\ {\small $r<0.25$}\end{tabular} & \begin{tabular}{@{}c@{}}{\small $\ell<500$} \\ {\small $r \geq 0.25$}\end{tabular} & \begin{tabular}{@{}c@{}}{\small $\ell>500$} \\ {\small $r<0.25$}\end{tabular} & \begin{tabular}{@{}c@{}}{\small $\ell>500$} \\ {\small $ r \geq 0.25$}\end{tabular} \\\hline
        {\small Baseline} & {\small 11.55} & {\small 62.41} & {\small 9.55} & {\small 76.08} \\ 
        {\small Oracle} & {\small 82.02} & {\small 93.89} & {\small 100.00} & {\small 100.00} \\ 
        {\small bert-base-cased} & {\small 41.60} & {\small 81.89} & {\small 48.48} & {\small 88.53} \\ 
        {\small distilbert-base-cased} & {\small 39.80} & {\small 81.32} & {\small 48.17} & {\small 88.50} \\ 
        {\small roberta-base} & {\small 32.87} & {\small 78.39} & {\small 45.16} & {\small 88.37} \\ 
        {\small distilroberta-base} & {\small 29.33} & {\small 76.48} & {\small 43.69} & {\small 86.86} \\ 
        {\small albert-base-v2} & {\small 40.44} & {\small 81.81} & {\small 51.58} & {\small 89.56} \\ 
        {\small xlnet-base-cased} & {\small 28.72} & {\small 75.07} & {\small 36.16} & {\small 84.29} \\ 
        {\small gpt2} & {\small 18.34} & {\small 69.56} & {\small 33.53} & {\small 82.74} \\ 
        \end{tabular}
        \caption{Performance for OntoNotes, no fine-tuning. Buckets respectively contain 6,949, 30,694, 1,649, and 11,123 query instances.}
        \label{tab:nofinetuningonto}
    \end{subtable}
    \hfill
    \begin{subtable}[h]{0.49\textwidth}
        \centering
        \begin{tabular}{c|rrrr}
        {\small Model} & \begin{tabular}{@{}c@{}}{\small $\ell<500$} \\ {\small $r<0.25$}\end{tabular} & \begin{tabular}{@{}c@{}}{\small $\ell<500$} \\ {\small $r \geq 0.25$}\end{tabular} & \begin{tabular}{@{}c@{}}{\small $\ell>500$} \\ {\small $r<0.25$}\end{tabular} & \begin{tabular}{@{}c@{}}{\small $\ell>500$} \\ {\small $ r \geq 0.25$}\end{tabular} \\\hline
        {\small Baseline} & {\small 11.56} & {\small 56.34} & {\small 18.25} & {\small 49.88} \\ 
        {\small Oracle} & {\small 96.41} & {\small 100.00} & {\small 100.00} & {\small 100.00} \\ 
        {\small bert-base-cased} & {\small 59.59} & {\small 83.54} & {\small 66.34} & {\small 89.39} \\ 
        {\small distilbert-base-cased} & {\small 58.06} & {\small 83.15} & {\small 65.09} & {\small 88.10} \\ 
        {\small roberta-base} & {\small 39.84} & {\small 76.79} & {\small 47.65} & {\small 80.01} \\ 
        {\small distilroberta-base} & {\small 32.42} & {\small 72.22} & {\small 42.29} & {\small 70.81} \\ 
        {\small albert-base-v2} & {\small 56.53} & {\small 82.22} & {\small 66.64} & {\small 88.44} \\ 
        {\small xlnet-base-cased} & {\small 35.76} & {\small 74.08} & {\small 37.68} & {\small 75.16} \\ 
        {\small gpt2} & {\small 21.75} & {\small 63.41} & {\small 33.33} & {\small 61.00} \\ 
        \end{tabular}
        \caption{Performance for PDEP, no fine-tuning. Buckets respectively contain 4,970, 1,618, 733, and 699 query instances.}
        \label{tab:nofinetuningpdep}
    \end{subtable}
    
    \vspace{10pt}
        
    \begin{subtable}[h]{0.49\textwidth}
        \begin{tabular}{c|rrrr}
        {\small Model} & \begin{tabular}{@{}c@{}}{\small $\ell<500$} \\ {\small $r<0.25$}\end{tabular} & \begin{tabular}{@{}c@{}}{\small $\ell<500$} \\ {\small $r \geq 0.25$}\end{tabular} & \begin{tabular}{@{}c@{}}{\small $\ell>500$} \\ {\small $r<0.25$}\end{tabular} & \begin{tabular}{@{}c@{}}{\small $\ell>500$} \\ {\small $ r \geq 0.25$}\end{tabular} \\\hline
        {\small Baseline} & {\small 11.55} & {\small 62.41} & {\small 9.55} & {\small 76.08} \\ 
        {\small Oracle} & {\small 82.02} & {\small 93.89} & {\small 100.00} & {\small 100.00} \\ 
        {\small bert-base-cased} & {\small 43.42} & {\small 82.37} & {\small 49.81} & {\small 89.45} \\ 
        {\small distilbert-base-cased} & {\small 41.62} & {\small 81.98} & {\small 50.31} & {\small 89.43} \\ 
        {\small roberta-base} & {\small 37.87} & {\small 80.68} & {\small 53.65} & {\small 89.43} \\ 
        {\small distilroberta-base} & {\small 34.74} & {\small 79.27} & {\small 48.50} & {\small 88.74} \\ 
        {\small albert-base-v2} & {\small 39.26} & {\small 81.50} & {\small 51.65} & {\small 89.31} \\ 
        {\small xlnet-base-cased} & {\small 37.53} & {\small 79.12} & {\small 51.40} & {\small 87.97} \\ 
        {\small gpt2} & {\small 18.12} & {\small 68.92} & {\small 32.99} & {\small 82.08} \\ 
        \end{tabular}
        \caption{Best performance across all fine-tuning trials for each model on OntoNotes.}
        \label{tab:finetuningonto}
    \end{subtable}
    \hfill
    \begin{subtable}[h]{0.49\textwidth}   
        \begin{tabular}{c|rrrr}
        {\small Model} & \begin{tabular}{@{}c@{}}{\small $\ell<500$} \\ {\small $r<0.25$}\end{tabular} & \begin{tabular}{@{}c@{}}{\small $\ell<500$} \\ {\small $r \geq 0.25$}\end{tabular} & \begin{tabular}{@{}c@{}}{\small $\ell>500$} \\ {\small $r<0.25$}\end{tabular} & \begin{tabular}{@{}c@{}}{\small $\ell>500$} \\ {\small $ r \geq 0.25$}\end{tabular} \\\hline
        {\small Baseline} & {\small 11.56} & {\small 56.34} & {\small 18.25} & {\small 49.88} \\ 
        {\small Oracle} & {\small 96.41} & {\small 100.00} & {\small 100.00} & {\small 100.00} \\ 
        {\small bert-base-cased} & {\small 60.37} & {\small 83.49} & {\small 67.87} & {\small 89.28} \\ 
        {\small distilbert-base-cased} & {\small 58.33} & {\small 82.91} & {\small 67.27} & {\small 87.52} \\ 
        {\small roberta-base} & {\small 48.99} & {\small 80.32} & {\small 60.04} & {\small 85.72} \\ 
        {\small distilroberta-base} & {\small 42.25} & {\small 77.25} & {\small 53.37} & {\small 79.19} \\ 
        {\small albert-base-v2} & {\small 53.75} & {\small 81.85} & {\small 65.57} & {\small 86.97} \\ 
        {\small xlnet-base-cased} & {\small 49.46} & {\small 80.50} & {\small 56.19} & {\small 84.53} \\ 
        {\small gpt2} & {\small 21.53} & {\small 63.00} & {\small 35.57} & {\small 61.08} \\ 
        \end{tabular}
        \caption{Best performance across all fine-tuning trials for each model on PDEP.}
        \label{tab:finetuningpdep}
    \end{subtable}
     
    \caption{
    Mean average precision performance broken down by corpus and model. Performance was further measured on different buckets of instances, as indicated in column headers: $\ell$ is a query instance's lemma frequency in $\mathcal{D}$, and $r$ is the proportional frequency of a query instance's sense across all instances of the lemma in $\mathcal{D}$.}
    \label{tab:maintable}
\end{table*}

\begin{figure}
    \centering
    \includegraphics[width=0.8\linewidth]{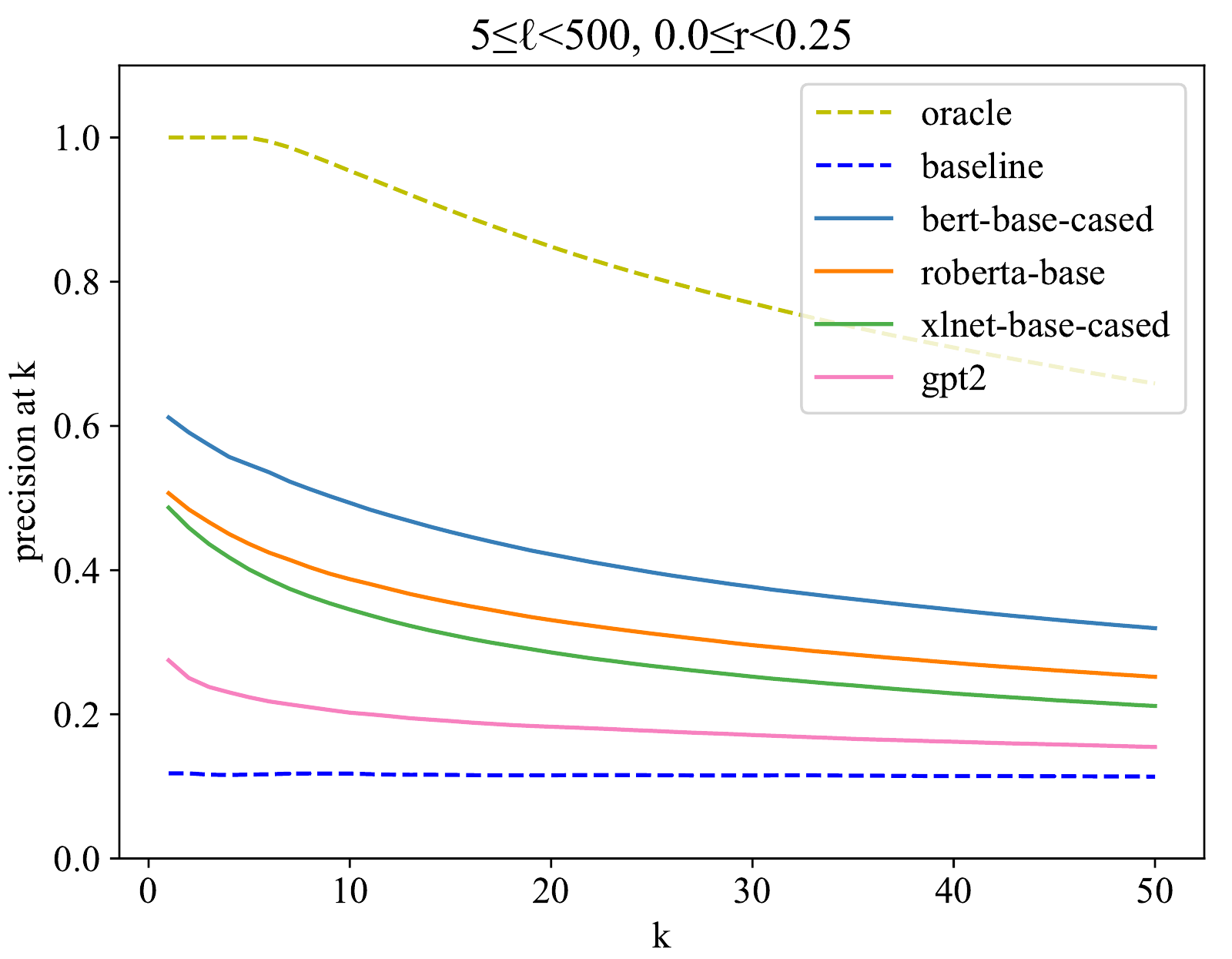}
    \caption{A sample of averaged precision at $k$ curves, showing performance of several models on OntoNotes for the bucket with lemmas occurring fewer than 500 times in $\mathcal{D}$ and senses with proportional frequencies lower than 0.25. Mean average precision, as shown in each cell in \Cref{tab:maintable}, is obtained by averaging every point along one of these lines.}
    \label{fig:patk}
\end{figure}

Main results are given in \Cref{tab:maintable}. 
Instances are bucketed by two parameters: $\ell$ is the instance's lemma's frequency in $\mathcal{D}$, and $r$ is the proportion of all instances of the lemma that have the same sense in $\mathcal{D}$.
Recall that each cell in \Cref{tab:maintable} is {\it mean} average precision over precision at k for 50 instances, and see \Cref{fig:patk} for a sample of what the P@K curves look like for each cell in \Cref{tab:maintable}.



Two example queries and top results appear in \Cref{fig:query_sample}.
Consider the results for the ``on'' query at right: 
out of a large haystack (1728 available tokens of ``on''), the system has correctly retrieved 6 of the 14 relevant needles, featuring 4 at the top of the list!
Interestingly, even though the complement of {\it on} in the query instance is just ``20'' \citep[a fused-head construction;][]{elazar-19} instead of the name of a drug or substance, the first four results are still relevant. 
From examining the top 50 results as well as the 8 false negatives (which include ``high on drugs'', ``drunk on Scotch'', etc.), 
it appears that BERT is prioritizing syntactic criteria (``on'' following a verb, especially a copula), possibly because the information following the query preposition is semantically impoverished. 
In fact, the top 3 results have a trigram match (``I was \textbf{on}'').

All models perform well above the random baseline across all buckets, and show little differentiation for buckets with only non-proportionally-rare senses $r \geq 0.25$.
Dramatic differences emerge, however, for buckets with proportionally rare senses $r<0.25$, with GPT-2 \citep{gpt2} showing the poorest performance.
In line with general trends, the distilled variants of models \citep{distilbert} generally track only a few points behind their undistilled variants.
Overall, {\tt\small bert-base-cased} performance is best. 
ALBERT \citep{albert} usually performs similarly, and RoBERTa \citep{roberta} and XLNet \citep{xlnet} show noticeably worse performance on rare senses.

It is especially surprising that RoBERTa performs so much worse than BERT on rare senses (cf. \Cref{tab:nofinetuningonto}, column 1), even though RoBERTa outperforms BERT on GLUE \citep{wang_glue_2018}, and RoBERTa is more closely related to BERT than XLNet or ALBERT: whereas XLNet and ALBERT differ architecturally from BERT, RoBERTa is architecturally identical and differs only in details of training. 
(For example, it implements dynamic masking, does not perform next sentence prediction, and uses an order of magnitude more data.)
Fine-tuning allows RoBERTa to close much of this gap but not all of it (cf.~\Cref{tab:finetuningonto}). 
This indicates that word sense information is less readily accessible in RoBERTa embeddings compared to BERT embeddings, and that fine-tuning cannot draw out this information to the extent that it already is in BERT.
Taken together, these results call into the question how conclusively a model can be judged based just on performance on downstream task benchmarks: RoBERTa performs better on GLUE, which has led many to prefer it as generally superior to BERT, but we have seen here it seems to be inferior within the domain of lexical semantics.

\section{Conclusion}

We have presented an evaluation method for probing the word sense content of contextualized word embeddings and applied it to several popular CWE models. 
We find that their ability to capture rare word senses using their representations is variable and not easily explained by each model's training and architectural characteristics.
Moreover, we find that performance of models on this word sense evaluation is not directly proportional to their performance on downstream task benchmarks like GLUE.
We view these differences between models as reason for further evaluation of CWE models to investigate their word sense representations, and inquiry into factors that affect word sense content of contextualized word embeddings.


%

\section*{Acknowledgments}

We thank Ken Litkowski for providing us with the PDEP corpus and for allowing us to distribute it with our code.

\bibliography{mypaper}
\bibliographystyle{acl_natbib}

\appendix

\section{Fine-tuning Setup}
\label{sec:ft}
For the fine-tuning step, a simple linear projection layer is added to the end of the CWE model. 
HuggingFace's implementation of AdamW \citep{loshchilov_decoupled_2019} is used as the optimizer with a learning rate of {\tt 2e-5} for 40 epochs.
Prepositions in STREUSLE are annotated with two supersenses, but only the first one is used for fine-tuning.

\clearpage

\section{Full Results, Main Experiment}
\label{sec:results_apdx}
{\tiny
\subsection{OntoNotes, $\ell<500$, $r<0.25$}
\begin{tabular}{lrrr}
Model & \# FT Instances & Precision & Recall \\
random baseline & 0 & 11.55 & 3.63 \\
oracle & 0 & 82.02 & 23.92 \\
roberta-base & 0 & 32.87 & 9.28 \\
roberta-base & 100 & 34.38 & 9.80 \\
roberta-base & 250 & 37.87 & 10.80 \\
roberta-base & 500 & 35.75 & 10.20 \\
roberta-base & 1000 & 36.88 & 10.48 \\
roberta-base & 2500 & 35.09 & 9.96 \\
bert-base-cased & 0 & 41.60 & 11.79 \\
bert-base-cased & 100 & 42.88 & 12.20 \\
bert-base-cased & 250 & 43.04 & 12.24 \\
bert-base-cased & 500 & 42.02 & 11.93 \\
bert-base-cased & 1000 & 43.42 & 12.37 \\
bert-base-cased & 2500 & 43.37 & 12.34 \\
distilroberta-base & 0 & 29.33 & 8.30 \\
distilroberta-base & 100 & 32.87 & 9.32 \\
distilroberta-base & 250 & 34.74 & 9.88 \\
distilroberta-base & 500 & 33.99 & 9.63 \\
distilroberta-base & 1000 & 32.24 & 9.16 \\
distilroberta-base & 2500 & 32.92 & 9.33 \\
distilbert-base-cased & 0 & 39.80 & 11.26 \\
distilbert-base-cased & 100 & 41.62 & 11.84 \\
distilbert-base-cased & 250 & 41.42 & 11.78 \\
distilbert-base-cased & 500 & 41.11 & 11.70 \\
distilbert-base-cased & 1000 & 40.70 & 11.55 \\
distilbert-base-cased & 2500 & 41.55 & 11.83 \\
gpt2 & 0 & 18.34 & 5.37 \\
gpt2 & 100 & 16.92 & 4.99 \\
gpt2 & 250 & 18.08 & 5.30 \\
gpt2 & 500 & 18.12 & 5.32 \\
gpt2 & 1000 & 17.59 & 5.18 \\
gpt2 & 2500 & 17.39 & 5.11 \\
albert-base-v2 & 0 & 40.44 & 11.50 \\
albert-base-v2 & 100 & 39.26 & 11.22 \\
albert-base-v2 & 250 & 37.24 & 10.66 \\
albert-base-v2 & 500 & 36.72 & 10.51 \\
albert-base-v2 & 1000 & 38.01 & 10.86 \\
albert-base-v2 & 2500 & 38.48 & 10.98 \\
xlnet-base-cased & 0 & 28.72 & 7.93 \\
xlnet-base-cased & 100 & 36.38 & 10.05 \\
xlnet-base-cased & 250 & 37.53 & 10.44 \\
xlnet-base-cased & 500 & 34.42 & 9.55 \\
xlnet-base-cased & 1000 & 37.38 & 10.36 \\
xlnet-base-cased & 2500 & 36.95 & 10.31 \\
\end{tabular}}

{\tiny
\subsection{OntoNotes, $\ell<500$, $r\geq 0.25$}
\begin{tabular}{lrrr}
Model & \# FT Instances & Precision & Recall \\
random baseline & 0 & 11.55 & 3.63 \\
oracle & 0 & 82.02 & 23.92 \\
roberta-base & 0 & 32.87 & 9.28 \\
roberta-base & 100 & 34.38 & 9.80 \\
roberta-base & 250 & 37.87 & 10.80 \\
roberta-base & 500 & 35.75 & 10.20 \\
roberta-base & 1000 & 36.88 & 10.48 \\
roberta-base & 2500 & 35.09 & 9.96 \\
bert-base-cased & 0 & 41.60 & 11.79 \\
bert-base-cased & 100 & 42.88 & 12.20 \\
bert-base-cased & 250 & 43.04 & 12.24 \\
bert-base-cased & 500 & 42.02 & 11.93 \\
bert-base-cased & 1000 & 43.42 & 12.37 \\
bert-base-cased & 2500 & 43.37 & 12.34 \\
distilroberta-base & 0 & 29.33 & 8.30 \\
distilroberta-base & 100 & 32.87 & 9.32 \\
distilroberta-base & 250 & 34.74 & 9.88 \\
distilroberta-base & 500 & 33.99 & 9.63 \\
distilroberta-base & 1000 & 32.24 & 9.16 \\
distilroberta-base & 2500 & 32.92 & 9.33 \\
distilbert-base-cased & 0 & 39.80 & 11.26 \\
distilbert-base-cased & 100 & 41.62 & 11.84 \\
distilbert-base-cased & 250 & 41.42 & 11.78 \\
distilbert-base-cased & 500 & 41.11 & 11.70 \\
distilbert-base-cased & 1000 & 40.70 & 11.55 \\
distilbert-base-cased & 2500 & 41.55 & 11.83 \\
gpt2 & 0 & 18.34 & 5.37 \\
gpt2 & 100 & 16.92 & 4.99 \\
gpt2 & 250 & 18.08 & 5.30 \\
gpt2 & 500 & 18.12 & 5.32 \\
gpt2 & 1000 & 17.59 & 5.18 \\
gpt2 & 2500 & 17.39 & 5.11 \\
albert-base-v2 & 0 & 40.44 & 11.50 \\
albert-base-v2 & 100 & 39.26 & 11.22 \\
albert-base-v2 & 250 & 37.24 & 10.66 \\
albert-base-v2 & 500 & 36.72 & 10.51 \\
albert-base-v2 & 1000 & 38.01 & 10.86 \\
albert-base-v2 & 2500 & 38.48 & 10.98 \\
xlnet-base-cased & 0 & 28.72 & 7.93 \\
xlnet-base-cased & 100 & 36.38 & 10.05 \\
xlnet-base-cased & 250 & 37.53 & 10.44 \\
xlnet-base-cased & 500 & 34.42 & 9.55 \\
xlnet-base-cased & 1000 & 37.38 & 10.36 \\
xlnet-base-cased & 2500 & 36.95 & 10.31 \\
\end{tabular}}

{\tiny
\subsection{OntoNotes, $\ell\geq 500$, $r<0.25$}
\begin{tabular}{lrrr}
Model & \# FT Instances & Precision & Recall \\
random baseline & 0 & 11.55 & 3.63 \\
oracle & 0 & 82.02 & 23.92 \\
roberta-base & 0 & 32.87 & 9.28 \\
roberta-base & 100 & 34.38 & 9.80 \\
roberta-base & 250 & 37.87 & 10.80 \\
roberta-base & 500 & 35.75 & 10.20 \\
roberta-base & 1000 & 36.88 & 10.48 \\
roberta-base & 2500 & 35.09 & 9.96 \\
bert-base-cased & 0 & 41.60 & 11.79 \\
bert-base-cased & 100 & 42.88 & 12.20 \\
bert-base-cased & 250 & 43.04 & 12.24 \\
bert-base-cased & 500 & 42.02 & 11.93 \\
bert-base-cased & 1000 & 43.42 & 12.37 \\
bert-base-cased & 2500 & 43.37 & 12.34 \\
distilroberta-base & 0 & 29.33 & 8.30 \\
distilroberta-base & 100 & 32.87 & 9.32 \\
distilroberta-base & 250 & 34.74 & 9.88 \\
distilroberta-base & 500 & 33.99 & 9.63 \\
distilroberta-base & 1000 & 32.24 & 9.16 \\
distilroberta-base & 2500 & 32.92 & 9.33 \\
distilbert-base-cased & 0 & 39.80 & 11.26 \\
distilbert-base-cased & 100 & 41.62 & 11.84 \\
distilbert-base-cased & 250 & 41.42 & 11.78 \\
distilbert-base-cased & 500 & 41.11 & 11.70 \\
distilbert-base-cased & 1000 & 40.70 & 11.55 \\
distilbert-base-cased & 2500 & 41.55 & 11.83 \\
gpt2 & 0 & 18.34 & 5.37 \\
gpt2 & 100 & 16.92 & 4.99 \\
gpt2 & 250 & 18.08 & 5.30 \\
gpt2 & 500 & 18.12 & 5.32 \\
gpt2 & 1000 & 17.59 & 5.18 \\
gpt2 & 2500 & 17.39 & 5.11 \\
albert-base-v2 & 0 & 40.44 & 11.50 \\
albert-base-v2 & 100 & 39.26 & 11.22 \\
albert-base-v2 & 250 & 37.24 & 10.66 \\
albert-base-v2 & 500 & 36.72 & 10.51 \\
albert-base-v2 & 1000 & 38.01 & 10.86 \\
albert-base-v2 & 2500 & 38.48 & 10.98 \\
xlnet-base-cased & 0 & 28.72 & 7.93 \\
xlnet-base-cased & 100 & 36.38 & 10.05 \\
xlnet-base-cased & 250 & 37.53 & 10.44 \\
xlnet-base-cased & 500 & 34.42 & 9.55 \\
xlnet-base-cased & 1000 & 37.38 & 10.36 \\
xlnet-base-cased & 2500 & 36.95 & 10.31 \\
\end{tabular}}

{\tiny
\subsection{OntoNotes, $\ell\geq 500$, $r\geq 0.25$}
\begin{tabular}{lrrr}
Model & \# FT Instances & Precision & Recall \\
random baseline & 0 & 11.55 & 3.63 \\
oracle & 0 & 82.02 & 23.92 \\
roberta-base & 0 & 32.87 & 9.28 \\
roberta-base & 100 & 34.38 & 9.80 \\
roberta-base & 250 & 37.87 & 10.80 \\
roberta-base & 500 & 35.75 & 10.20 \\
roberta-base & 1000 & 36.88 & 10.48 \\
roberta-base & 2500 & 35.09 & 9.96 \\
bert-base-cased & 0 & 41.60 & 11.79 \\
bert-base-cased & 100 & 42.88 & 12.20 \\
bert-base-cased & 250 & 43.04 & 12.24 \\
bert-base-cased & 500 & 42.02 & 11.93 \\
bert-base-cased & 1000 & 43.42 & 12.37 \\
bert-base-cased & 2500 & 43.37 & 12.34 \\
distilroberta-base & 0 & 29.33 & 8.30 \\
distilroberta-base & 100 & 32.87 & 9.32 \\
distilroberta-base & 250 & 34.74 & 9.88 \\
distilroberta-base & 500 & 33.99 & 9.63 \\
distilroberta-base & 1000 & 32.24 & 9.16 \\
distilroberta-base & 2500 & 32.92 & 9.33 \\
distilbert-base-cased & 0 & 39.80 & 11.26 \\
distilbert-base-cased & 100 & 41.62 & 11.84 \\
distilbert-base-cased & 250 & 41.42 & 11.78 \\
distilbert-base-cased & 500 & 41.11 & 11.70 \\
distilbert-base-cased & 1000 & 40.70 & 11.55 \\
distilbert-base-cased & 2500 & 41.55 & 11.83 \\
gpt2 & 0 & 18.34 & 5.37 \\
gpt2 & 100 & 16.92 & 4.99 \\
gpt2 & 250 & 18.08 & 5.30 \\
gpt2 & 500 & 18.12 & 5.32 \\
gpt2 & 1000 & 17.59 & 5.18 \\
gpt2 & 2500 & 17.39 & 5.11 \\
albert-base-v2 & 0 & 40.44 & 11.50 \\
albert-base-v2 & 100 & 39.26 & 11.22 \\
albert-base-v2 & 250 & 37.24 & 10.66 \\
albert-base-v2 & 500 & 36.72 & 10.51 \\
albert-base-v2 & 1000 & 38.01 & 10.86 \\
albert-base-v2 & 2500 & 38.48 & 10.98 \\
xlnet-base-cased & 0 & 28.72 & 7.93 \\
xlnet-base-cased & 100 & 36.38 & 10.05 \\
xlnet-base-cased & 250 & 37.53 & 10.44 \\
xlnet-base-cased & 500 & 34.42 & 9.55 \\
xlnet-base-cased & 1000 & 37.38 & 10.36 \\
xlnet-base-cased & 2500 & 36.95 & 10.31 \\
\end{tabular}}

{\tiny
\subsection{PDEP, $\ell<500$, $r<0.25$}
\begin{tabular}{lrrr}
Model & \# FT Instances & Precision & Recall \\
random baseline & 0 & 11.56 & 1.85 \\
oracle & 0 & 96.41 & 15.19 \\
roberta-base & 0 & 39.84 & 5.90 \\
roberta-base & 100 & 42.09 & 6.29 \\
roberta-base & 250 & 47.86 & 7.21 \\
roberta-base & 500 & 45.20 & 6.78 \\
roberta-base & 1000 & 48.99 & 7.37 \\
roberta-base & 2500 & 46.63 & 6.99 \\
bert-base-cased & 0 & 59.59 & 8.99 \\
bert-base-cased & 100 & 60.00 & 9.07 \\
bert-base-cased & 250 & 60.23 & 9.11 \\
bert-base-cased & 500 & 60.07 & 9.07 \\
bert-base-cased & 1000 & 59.94 & 9.07 \\
bert-base-cased & 2500 & 60.37 & 9.13 \\
distilroberta-base & 0 & 32.42 & 4.80 \\
distilroberta-base & 100 & 39.20 & 5.84 \\
distilroberta-base & 250 & 42.25 & 6.34 \\
distilroberta-base & 500 & 41.75 & 6.26 \\
distilroberta-base & 1000 & 40.09 & 6.02 \\
distilroberta-base & 2500 & 39.89 & 5.96 \\
distilbert-base-cased & 0 & 58.06 & 8.75 \\
distilbert-base-cased & 100 & 58.06 & 8.77 \\
distilbert-base-cased & 250 & 58.02 & 8.76 \\
distilbert-base-cased & 500 & 57.96 & 8.75 \\
distilbert-base-cased & 1000 & 58.03 & 8.75 \\
distilbert-base-cased & 2500 & 58.33 & 8.79 \\
gpt2 & 0 & 21.75 & 3.24 \\
gpt2 & 100 & 18.34 & 2.76 \\
gpt2 & 250 & 21.53 & 3.17 \\
gpt2 & 500 & 19.77 & 2.93 \\
gpt2 & 1000 & 19.42 & 2.90 \\
gpt2 & 2500 & 20.58 & 3.08 \\
albert-base-v2 & 0 & 56.53 & 8.55 \\
albert-base-v2 & 100 & 53.75 & 8.14 \\
albert-base-v2 & 250 & 52.56 & 7.95 \\
albert-base-v2 & 500 & 51.63 & 7.80 \\
albert-base-v2 & 1000 & 51.97 & 7.86 \\
albert-base-v2 & 2500 & 53.10 & 8.03 \\
xlnet-base-cased & 0 & 35.76 & 5.12 \\
xlnet-base-cased & 100 & 48.28 & 7.02 \\
xlnet-base-cased & 250 & 48.47 & 7.08 \\
xlnet-base-cased & 500 & 45.99 & 6.68 \\
xlnet-base-cased & 1000 & 48.80 & 7.13 \\
xlnet-base-cased & 2500 & 49.46 & 7.25 \\
\end{tabular}}

{\tiny
\subsection{PDEP, $\ell<500$, $r\geq 0.25$}
\begin{tabular}{lrrr}
Model & \# FT Instances & Precision & Recall \\
random baseline & 0 & 11.56 & 1.85 \\
oracle & 0 & 96.41 & 15.19 \\
roberta-base & 0 & 39.84 & 5.90 \\
roberta-base & 100 & 42.09 & 6.29 \\
roberta-base & 250 & 47.86 & 7.21 \\
roberta-base & 500 & 45.20 & 6.78 \\
roberta-base & 1000 & 48.99 & 7.37 \\
roberta-base & 2500 & 46.63 & 6.99 \\
bert-base-cased & 0 & 59.59 & 8.99 \\
bert-base-cased & 100 & 60.00 & 9.07 \\
bert-base-cased & 250 & 60.23 & 9.11 \\
bert-base-cased & 500 & 60.07 & 9.07 \\
bert-base-cased & 1000 & 59.94 & 9.07 \\
bert-base-cased & 2500 & 60.37 & 9.13 \\
distilroberta-base & 0 & 32.42 & 4.80 \\
distilroberta-base & 100 & 39.20 & 5.84 \\
distilroberta-base & 250 & 42.25 & 6.34 \\
distilroberta-base & 500 & 41.75 & 6.26 \\
distilroberta-base & 1000 & 40.09 & 6.02 \\
distilroberta-base & 2500 & 39.89 & 5.96 \\
distilbert-base-cased & 0 & 58.06 & 8.75 \\
distilbert-base-cased & 100 & 58.06 & 8.77 \\
distilbert-base-cased & 250 & 58.02 & 8.76 \\
distilbert-base-cased & 500 & 57.96 & 8.75 \\
distilbert-base-cased & 1000 & 58.03 & 8.75 \\
distilbert-base-cased & 2500 & 58.33 & 8.79 \\
gpt2 & 0 & 21.75 & 3.24 \\
gpt2 & 100 & 18.34 & 2.76 \\
gpt2 & 250 & 21.53 & 3.17 \\
gpt2 & 500 & 19.77 & 2.93 \\
gpt2 & 1000 & 19.42 & 2.90 \\
gpt2 & 2500 & 20.58 & 3.08 \\
albert-base-v2 & 0 & 56.53 & 8.55 \\
albert-base-v2 & 100 & 53.75 & 8.14 \\
albert-base-v2 & 250 & 52.56 & 7.95 \\
albert-base-v2 & 500 & 51.63 & 7.80 \\
albert-base-v2 & 1000 & 51.97 & 7.86 \\
albert-base-v2 & 2500 & 53.10 & 8.03 \\
xlnet-base-cased & 0 & 35.76 & 5.12 \\
xlnet-base-cased & 100 & 48.28 & 7.02 \\
xlnet-base-cased & 250 & 48.47 & 7.08 \\
xlnet-base-cased & 500 & 45.99 & 6.68 \\
xlnet-base-cased & 1000 & 48.80 & 7.13 \\
xlnet-base-cased & 2500 & 49.46 & 7.25 \\
\end{tabular}}

{\tiny
\subsection{PDEP, $\ell\geq 500$, $r<0.25$}
\begin{tabular}{lrrr}
Model & \# FT Instances & Precision & Recall \\
random baseline & 0 & 11.56 & 1.85 \\
oracle & 0 & 96.41 & 15.19 \\
roberta-base & 0 & 39.84 & 5.90 \\
roberta-base & 100 & 42.09 & 6.29 \\
roberta-base & 250 & 47.86 & 7.21 \\
roberta-base & 500 & 45.20 & 6.78 \\
roberta-base & 1000 & 48.99 & 7.37 \\
roberta-base & 2500 & 46.63 & 6.99 \\
bert-base-cased & 0 & 59.59 & 8.99 \\
bert-base-cased & 100 & 60.00 & 9.07 \\
bert-base-cased & 250 & 60.23 & 9.11 \\
bert-base-cased & 500 & 60.07 & 9.07 \\
bert-base-cased & 1000 & 59.94 & 9.07 \\
bert-base-cased & 2500 & 60.37 & 9.13 \\
distilroberta-base & 0 & 32.42 & 4.80 \\
distilroberta-base & 100 & 39.20 & 5.84 \\
distilroberta-base & 250 & 42.25 & 6.34 \\
distilroberta-base & 500 & 41.75 & 6.26 \\
distilroberta-base & 1000 & 40.09 & 6.02 \\
distilroberta-base & 2500 & 39.89 & 5.96 \\
distilbert-base-cased & 0 & 58.06 & 8.75 \\
distilbert-base-cased & 100 & 58.06 & 8.77 \\
distilbert-base-cased & 250 & 58.02 & 8.76 \\
distilbert-base-cased & 500 & 57.96 & 8.75 \\
distilbert-base-cased & 1000 & 58.03 & 8.75 \\
distilbert-base-cased & 2500 & 58.33 & 8.79 \\
gpt2 & 0 & 21.75 & 3.24 \\
gpt2 & 100 & 18.34 & 2.76 \\
gpt2 & 250 & 21.53 & 3.17 \\
gpt2 & 500 & 19.77 & 2.93 \\
gpt2 & 1000 & 19.42 & 2.90 \\
gpt2 & 2500 & 20.58 & 3.08 \\
albert-base-v2 & 0 & 56.53 & 8.55 \\
albert-base-v2 & 100 & 53.75 & 8.14 \\
albert-base-v2 & 250 & 52.56 & 7.95 \\
albert-base-v2 & 500 & 51.63 & 7.80 \\
albert-base-v2 & 1000 & 51.97 & 7.86 \\
albert-base-v2 & 2500 & 53.10 & 8.03 \\
xlnet-base-cased & 0 & 35.76 & 5.12 \\
xlnet-base-cased & 100 & 48.28 & 7.02 \\
xlnet-base-cased & 250 & 48.47 & 7.08 \\
xlnet-base-cased & 500 & 45.99 & 6.68 \\
xlnet-base-cased & 1000 & 48.80 & 7.13 \\
xlnet-base-cased & 2500 & 49.46 & 7.25 \\
\end{tabular}}

{\tiny
\subsection{PDEP, $\ell\geq 500$, $r\geq 0.25$}
\begin{tabular}{lrrr}
Model & \# FT Instances & Precision & Recall \\
random baseline & 0 & 11.56 & 1.85 \\
oracle & 0 & 96.41 & 15.19 \\
roberta-base & 0 & 39.84 & 5.90 \\
roberta-base & 100 & 42.09 & 6.29 \\
roberta-base & 250 & 47.86 & 7.21 \\
roberta-base & 500 & 45.20 & 6.78 \\
roberta-base & 1000 & 48.99 & 7.37 \\
roberta-base & 2500 & 46.63 & 6.99 \\
bert-base-cased & 0 & 59.59 & 8.99 \\
bert-base-cased & 100 & 60.00 & 9.07 \\
bert-base-cased & 250 & 60.23 & 9.11 \\
bert-base-cased & 500 & 60.07 & 9.07 \\
bert-base-cased & 1000 & 59.94 & 9.07 \\
bert-base-cased & 2500 & 60.37 & 9.13 \\
distilroberta-base & 0 & 32.42 & 4.80 \\
distilroberta-base & 100 & 39.20 & 5.84 \\
distilroberta-base & 250 & 42.25 & 6.34 \\
distilroberta-base & 500 & 41.75 & 6.26 \\
distilroberta-base & 1000 & 40.09 & 6.02 \\
distilroberta-base & 2500 & 39.89 & 5.96 \\
distilbert-base-cased & 0 & 58.06 & 8.75 \\
distilbert-base-cased & 100 & 58.06 & 8.77 \\
distilbert-base-cased & 250 & 58.02 & 8.76 \\
distilbert-base-cased & 500 & 57.96 & 8.75 \\
distilbert-base-cased & 1000 & 58.03 & 8.75 \\
distilbert-base-cased & 2500 & 58.33 & 8.79 \\
gpt2 & 0 & 21.75 & 3.24 \\
gpt2 & 100 & 18.34 & 2.76 \\
gpt2 & 250 & 21.53 & 3.17 \\
gpt2 & 500 & 19.77 & 2.93 \\
gpt2 & 1000 & 19.42 & 2.90 \\
gpt2 & 2500 & 20.58 & 3.08 \\
albert-base-v2 & 0 & 56.53 & 8.55 \\
albert-base-v2 & 100 & 53.75 & 8.14 \\
albert-base-v2 & 250 & 52.56 & 7.95 \\
albert-base-v2 & 500 & 51.63 & 7.80 \\
albert-base-v2 & 1000 & 51.97 & 7.86 \\
albert-base-v2 & 2500 & 53.10 & 8.03 \\
xlnet-base-cased & 0 & 35.76 & 5.12 \\
xlnet-base-cased & 100 & 48.28 & 7.02 \\
xlnet-base-cased & 250 & 48.47 & 7.08 \\
xlnet-base-cased & 500 & 45.99 & 6.68 \\
xlnet-base-cased & 1000 & 48.80 & 7.13 \\
xlnet-base-cased & 2500 & 49.46 & 7.25 \\
\end{tabular}}

\end{document}